\documentclass[twocolumn]{autart}    %
\usepackage{graphicx}       
\usepackage{amsmath,amssymb}
\usepackage{cite}
\usepackage{nicefrac}
\usepackage{cases}
\usepackage{amsthm}
\usepackage{algorithm}

\let\classAND\AND
\let\AND\relax

\usepackage{algorithmic}

\let\AND\classAND

\AtBeginEnvironment{algorithmic}{\let\AND\algoAND}

\usepackage{hyperref}
\hypersetup{
    colorlinks = true,
    citecolor = {blue},
    linkcolor=blue
}
\edef\endfrontmatter{%
  \unexpanded\expandafter{\endfrontmatter}%
  \noexpand\endNoHyper %
}
\allowdisplaybreaks

\newtheorem{theorem}{Theorem}
\newtheorem{lemma}{Lemma}

\newtheorem{corollary}{Corollary}
\newtheorem{remark}{Remark}
\newtheorem*{assumption*}{\assumptionnumber}
\providecommand{\assumptionnumber}{}
\makeatletter
\newenvironment{assumption}[1]
{%
	\renewcommand{\assumptionnumber}{A#1}%
	\begin{assumption*}%
		\protected@edef\@currentlabel{A#1}%
	}
	{%
	\end{assumption*}
}
\makeatother

\usepackage{xpatch}
\makeatletter %
\xpatchcmd{\runningauthor@fmt}{\global\edef}{\protected@xdef}{}{}
\xpatchcmd{\runningauthor@fmt}{\global\edef}{\protected@xdef}{}{}
\xpatchcmd{\author@fmt}{\edef}{\protected@edef}{}{}
\def\@xnamedef#1{\expandafter\protected@xdef\csname #1\endcsname}
\def\ead@au#1{\protected@edef\@ead@au{#1}}
\def\add@xtok#1#2{\begingroup
  \protected@xdef\@act{\global\noexpand#1{\the#1#2}}\@act
\endgroup}
\def\no@harm{}
\makeatother

\def \newtext{\color{black}}
\def \roundtwo{\color{black}}

\newcommand{\tr }{^\mathrm{T}}
\newcommand{\ntr}{^\mathrm{-T}}

\newcommand{\abs}[1]{\left\lvert#1\right\rvert}
\newcommand{\norm}[1]{\left\lVert#1\right\rVert}
\newcommand{\indnorm}[1]{{\left\vert\kern-0.25ex\left\vert\kern-0.25ex\left\vert #1 
		\right\vert\kern-0.25ex\right\vert\kern-0.25ex\right\vert}}
\newcommand{\defeq}{\doteq}
\newcommand{\confreg}[1][2]{\mathcal{C}_{#1}}

\newcommand{\BR}{\mathbb{R}}
\newcommand{\BE}{\mathbb{E}}
\newcommand{\BP}{\mathbb{P}}
\newcommand{\CA}{\mathcal{A}}
\newcommand{\CB}{\mathcal{B}}
\newcommand{\CG}{\mathcal{G}}

\begin{document}

\begin{frontmatter}
\title{Sample Complexity of the Sign-Perturbed Sums Method\!\thanksref{footnoteinfo}} %

\thanks[footnoteinfo]{This research was supported by the European Union within the framework of the National Laboratory for Autonomous Systems, RRF-2.3.1-21-2022-00002; and by the TKP2021-NKTA-01 grant of the National Research, Development and Innovation Office (NRDIO), Hungary.}

\author[sztaki]{Szabolcs Szentp\'eteri}\ead{szentpes@sztaki.hu}\qquad
\author[sztaki,elte]{Bal\'azs Csan\'ad Cs\'aji}\ead{csaji@sztaki.hu}

\address[sztaki]{Institute for Computer Science and Control (SZTAKI), Hungarian Research Network (HUN-REN), Budapest, Hungary}
\address[elte]{Department of Probability Theory and Statistics, Institute of Mathematics, E\"otv\"os Lor\'and University (ELTE), Hungary}

\begin{keyword}                           %
linear regression, sample complexity, least squares method, confidence regions, finite sample properties
\end{keyword}                             %

\begin{abstract}                          %
We study the sample complexity of the Sign-Perturbed Sums (SPS) method, which constructs exact, non-asymptotic confidence regions for the true system parameters under mild statistical assumptions, such as independent and symmetric noise terms. The standard version of SPS deals with linear regression problems, however, it can be generalized to stochastic linear (dynamical) systems, even with closed-loop setups, and to nonlinear and nonparametric problems, as well. Although the strong consistency of the method was rigorously proven, the sample complexity of the algorithm was only analyzed so far for scalar linear regression problems. In this paper we study the sample complexity of SPS for general linear regression problems. We establish high probability upper bounds for the diameters of SPS confidence regions for finite sample sizes and show that the SPS regions shrink at the same, optimal rate as the classical asymptotic confidence ellipsoids. Finally, the difference between the theoretical bounds and the empirical sizes of SPS confidence regions is investigated experimentally. 
\end{abstract}

\end{frontmatter}

\setlength{\abovedisplayskip}{3mm}
\setlength{\belowdisplayskip}{3mm}

\section{Introduction}\label{sec:introduction}
\vspace{-1mm}
{\newtext System identification studies the problem of constructing mathematical models 
from empirical data, which is also important for several other fields, such as 
machine learning and statistics.} While classical results in the aforementioned
areas mainly focus on {\em asymptotic} properties and guarantees \cite{Ljung1999}, in recent years significant emphasis {\newtext has been given} to {\em non-asymptotic} approaches \cite{Ziemann2023}.
Particularly, lately both the control and machine learning communities gave considerable attention to study the finite-sample behaviour of {\em stochastic linear systems}.

One of the most widely used 
methods for linear regression is the {\em least squares} (LS) estimator. It is well-known that the LS estimator is the {\em best linear unbiased estimator} (BLUE), for example, assuming uncorrelated, {\newtext homoscedastic} noises, and it is {\em asymptotically efficient} under mild conditions, i.e., its asymptotic covariance matrix reaches the Cramér-Rao lower bound. Furthermore, the LS error decreases at the {\em optimal} rate of $\mathcal{O}(1/\sqrt{n})$, 
where $n$ is the sample size.

The (non-asymptotic) {\em sample complexity} of 
LS regression in case of {\em bounded} noises was studied in \cite{shamir2015sample}. The author established an $\mathcal{O}(1/\sqrt{n})$ lower bound on the 
regret,
and showed that it matches the upper bound up to logarithmic factors. {\em Probably approximately correct} (PAC) type upper bounds for the LS estimation 
error are also investigated in \cite{krikheli2021finite}, where it was shown that similar sample complexity bounds, with $\mathcal{O}(1/\sqrt{n})$ rate, hold
in the case of {\em subgaussian} noises.

In recent years the LS 
scheme was also analysed in the closed-loop linear system identification setting. 
Some of these recent works study the non-asymptotic properties of closed-loop linear system identification under {\em strong statistical assumptions} on the noises, such as  {\em joint Gaussianity}. In \cite{simchowitz18a}, PAC bounds for the estimation error of a stable transition matrix in an observable state-space setting was investigated. In the non-observable regime, finite sample bounds for the estimation error of a Hankel-type matrix was studied in \cite{sarkar2021}, while \cite{oymak2021} investigated the same properties in case of learning the Markov parameters of the system.
As most of these techniques assume that the noises and disturbances follow specific distributions, {\em distribution-free} (as well as non-asymptotic) techniques 
{\newtext still remain an important area of research \cite{Algo2018}}.

Two important examples of system identification algorithms that can construct 
finite sample confidence regions for the true parameter 
for any 
sample size, in a distribution-free setting, are the LSCR: {\em Leave-out Sign-dominant Correlation Regions} \cite{Campi2005} and the SPS: {\em Sign-Perturbed Sums} \cite{Csaji2015} methods. Here we focus on SPS.

Standard SPS assumes linear regression problems and constructs {\em exact} confidence regions around the LS estimate for any finite sample size under mild assumptions on the noises, namely that they are independent and their 
 distributions are {\em symmetric} about zero \cite{Csaji2012b}.

SPS was generalized to general linear systems, even for closed-loop settings \cite{Csaji2015cdc}.
An instrumental variables (IV) based extensions of SPS was investigated in \cite{volpe2015sign} for ARX systems, and in \cite{szentpeteri2023} for state-space models even under feedback. In \cite{kieffer2014guaranteed} a guaranteed characterization of SPS was developed using interval analysis, while in \cite{Csaji2015}
an ellipsoidal outer approximation was given, in case of linear regression. Several extensions of SPS were suggested, such as Data Perturbation (DP) methods \cite{kolumban2015perturbed} which can be used with other perturbations, not only sign changes{\newtext, and SPS was also combined with kernels and Bayesian inference \cite{baggio2022bayesian} leading to Bayesian frequentist bounds.}

Theoretical properties of SPS, such as its {\em exact coverage} probability \cite{Csaji2015} and its {\em strong consistency} \cite{Weyer2017} were proven for linear regression problems. %
Although {\em asymptotic} guarantees hold for the sizes of the SPS regions \cite{Weyer2017}, the finite sample properties have not been thoroughly investigated. Our first result regarding the sample complexity of SPS was presented in \cite{szentpeteri2023scalar}, where we established {\em geometric bounds} for the length of SPS confidence intervals for the case of {\em scalar} linear regression problems.

In this work we prove high probability upper bounds for the sizes of the SPS confidence regions for {\em general} linear regression, which are rigorous for {\em finite} samples. Our 
{\em non-asymptotic} bounds have the same shrinkage rate as that of the asymptotic confidence ellipsoids.
While most standard sample complexity analysis investigate the finite sample behaviour of some estimation error, our results provide a technical analysis of a {\em data-driven}, {\em distribution-free} confidence region construction.

The {\em main contributions} of the paper are as follows:
\vspace{-1mm}
\begin{enumerate}
    \item High probability upper bounds are derived for the sizes of SPS confidence regions for general linear regression, assuming subgaussian noises. {\newtext Our bounds are non-asymptotic
    and they show that the shrinkage rate of SPS is optimal.}\\[-0.5mm]
    \item Simulation experiments demonstrating the difference between the obtained theoretical bounds and the empirical performance are also shown.
\end{enumerate}
The paper is organized as follows. In Section \ref{sec:problem_setting} the problem setting and our main assumptions are introduced. Section \ref{sec:SPS_overview} gives a summary of the SPS algorithm. In Section \ref{sec:main_result} a theorem regarding the sample complexity of the SPS-Indicator algorithm is proved. The simulation experiment and comparisons are presented in Section \ref{sec:experiments}. Finally, Section \ref{sec:conclusion} summarizes and concludes the paper.

\section{Problem Setting}\label{sec:problem_setting}
\vspace{-1mm}
This section formalizes the addressed linear regression problem and introduces our main assumptions.
\subsection{Data Generation}
Consider the following linear regression problem
\begin{equation}\label{equ:system}
    Y_t \,\defeq\, \varphi_t\tr \theta^* + W_t,
\end{equation}
where $\varphi_t$ is a $d$-dimensional deterministic regressor, $Y_t$ is the scalar output, $W_t$ is the (random) scalar noise and $\theta^*$ is the $d$-dimensional (constant) true parameter to be estimated. We are given a sample of size $n$ which consists of $\varphi_1, \dots, \varphi_n$ (inputs) and $Y_1, \dots, Y_n$ (outputs).

Throughout the paper we will use the following notation:
\begin{align}
    &\Phi_n \defeq \begin{bmatrix}
        \varphi_1\tr \\
        \varphi_2\tr \\
        \vdots\\
        \varphi_n\tr 
    \end{bmatrix},\qquad
    w_n \defeq \begin{bmatrix}
        W_1\\
        W_2\\
        \vdots\\
        W_n
    \end{bmatrix},\qquad
    y_n \defeq \begin{bmatrix}
        Y_1\\
        Y_2\\
        \vdots\\
        Y_n
    \end{bmatrix},\\
    &	R_n \defeq \sum_{t=1}^{n}\varphi_t\varphi_t\tr  = \Phi_n\tr  \Phi_n,\\
    &	\bar{R}_n \defeq \frac{1}{n}\sum_{t=1}^{n}\varphi_t\varphi_t\tr  = \frac{1}{n}\Phi_n\tr  \Phi_n = \frac{1}{n}R_n.
\end{align}
Note that in our work we consider {\em deterministic} regressors $\{\varphi_t\}$; however, our results can be easily generalized to random exogenous regressors, where the regressor sequence $\{\varphi_t\}$ is independent of the noise sequence $\{W_t\}$. In that case, our assumptions on the regressors (stated below) must be satisfied almost surely, and then the analysis can be traced back to the presented theory by fixing a realization of the regressors (i.e., by conditioning on the $\sigma$-algebra generated by the regressors)
and applying the presented results realization-wise, see also \cite{Csaji2015}.
\subsection{Assumptions}
Our main assumptions are as follows.\\
\begin{assumption}{1}\label{assu:noise}
The noise sequence $\{W_t\}$ is independent and contains nonatomic, $\sigma$-subgaussian random variables whose probability distributions are symmetric about zero. 
\end{assumption}
\vspace{-1mm}

Recall that we call a random variable $W$ $\sigma${\em-subgaussian}, if 
for all $\lambda \in \mathbb{R},$ its moment generating function satisfies
    \begin{align}
    \mathbb{E}\big[\exp(\lambda W)\big]\, \leq\, \exp\!{\left(\frac{\lambda^2\sigma^2}{2}\right)}.
    \end{align}
Standard examples of subgaussian distributions are the Gaussian, and {\em any} distribution with a bounded support, such as the uniform, triangular and beta distributions.

Note that random variable $W$ is called {\em symmetric} about zero if\, $W$ has the same probability distribution as $-W$.
Moreover, we call random variable $W$ {\em nonatomic} if, for all constant $w \in \mathbb{R}$, we have $\mathbb{P}(W = w) = 0.$ Naturally, 
every continuous probability distribution is nonatomic.
Our assumptions on the noises are rather weak, e.g., their distributions can change over time and 
a wide range of distributions are subgaussian. Moreover, SPS itself does not exploit subgaussianity, it is only needed for the sample complexity analysis.
{\newtext The nonatomicity
of the distributions is nonessential, it is just used to avoid ties, see Algorithm \ref{alg:sps_indicator}, and hence to simplify the analysis.}\\
\begin{assumption}{2}\label{assu:R_nonsigular}
The regressor vectors are ``completely exciting'' in the sense that any $d$ regressors span the whole space, $\BR^d$. 
\end{assumption}
\vspace{-1mm}

Hence, for any subset\, $T$ of the index set $[n] \defeq \{1, \dots, n\}$ with $|\hspace{0.4mm}T\hspace{0.2mm}|=d$, i.e., having cardinality $d$, it holds that
\vspace{-1mm}
\begin{equation}
    \det\!\left(\frac{1}{d}\sum_{t \in T}\varphi_t\varphi_t\tr \right)\,\neq\, 0.
    \vspace{-3mm}
\end{equation}

\ref{assu:R_nonsigular} ensures the {\em boundedness} of SPS confidence regions, under suitable assumptions on the perturbations \cite{Care2022}.\\
\begin{assumption}{3}\label{assu:R_min_eigval}
The excitations are nonvanishing in the sense that there exists {\roundtwo a} constant\, $\lambda_0 > 0$, such that for all\, $n \geq d:$
\begin{align}
     \lambda_{\text{min}}(\bar{R}_n)\, \geq\, \lambda_0\, > \,0,
\end{align}
where\, $\lambda_{\text{min}}$ denotes the smallest eigenvalue.
\end{assumption}
\vspace{-3mm}
Assumption \ref{assu:R_min_eigval} guarantees that the averaged ``magnitude'' of the excitation does not get too small, 
which can provide a lower bound on the signal-to-noise ratio.\\
\begin{assumption}{4}\label{assu:Phi_Q_coherence}
Let\, $\Phi_n = \Phi_{{\scriptscriptstyle Q},n}\Phi_{{\scriptscriptstyle R},n}$ be the thin QR-decomposition of\, $\Phi_n$. 
There are constants $\kappa >0$ and\, $0 <\rho \leq 1$, such that the following upper bound holds for all\, $n \geq d:$
\begin{align}
    \mu(\Phi_n)\, \defeq\, \frac{n}{d} \max_{1\leq i \leq n}\norm{\Phi_{{\scriptscriptstyle Q},n}\tr e_i}^2 \leq\, \kappa\, n^{1-\rho},
\end{align}
 where $\mu(\Phi_{n})$ is called the coherence of\, $\Phi_{n}$.
\end{assumption}
In assumption \ref{assu:Phi_Q_coherence} we used the definition of coherence from \cite[Definition 1.2]{Candes2008ExactMC} and the facts that $\text{range}(\Phi_n) = \text{range}(\Phi_{{\scriptscriptstyle Q},n})$ and $\Phi_{{\scriptscriptstyle Q},n}\Phi_{{\scriptscriptstyle Q},n}\tr$ is an orthogonal projection onto $\text{range}(\Phi_n)$.
Note that from this definition of coherence, it follows that $1 \leq \mu(\Phi_n) \leq n/d$ \cite{Candes2008ExactMC}.
 Our coherence assumption 
 ensures that the excitation does not ``concentrate'' too much 
 to any specific regressor.
 Throughout the paper we will assume \ref{assu:Phi_Q_coherence} alongside \ref{assu:R_nonsigular}, therefore, the thin QR-decomposition of $\Phi_n = \Phi_{{\scriptscriptstyle Q},n}\Phi_{{\scriptscriptstyle R},n}$ is unique, since matrix $\Phi_{{\scriptscriptstyle R},n}$ is full rank. 
 
 Intuitively, assumption \ref{assu:R_nonsigular} ensures that the regressors have ``sufficiently diverse'' directions; assumption \ref{assu:R_min_eigval} guarantees a lower bound on the ``energy'' of the excitation; and, finally, assumption \ref{assu:Phi_Q_coherence} ensures that the excitation is not too ``unevenly distributed'' 
 among the regressors, measured by the ``coherence'' parameters.

The SPS algorithm, detailed in Section \ref{sec:SPS_overview}, requires much milder assumptions on the noises and on the regressors. The stronger assumptions are needed to give {\em non-asymptotic} results on the {\em sample complexity} of SPS, for {\em {\newtext almost all} realizations} of the regressor vectors.

 {\newtext A typical choice of regressors that satisfy assumptions \ref{assu:R_nonsigular}-\ref{assu:Phi_Q_coherence} could be (almost all realizations of) i.i.d. continuous random vectors with a positive definite covariance matrix. In this case,  \ref{assu:R_nonsigular} almost surely (a.s.) holds,  since the distribution is continuous and it does not concentrate to any proper affine subspace (which is a consequence of having a positive definite covariance matrix). Furthermore, because the regressors are i.i.d., the excitations (a.s.) do not vanish (\ref{assu:R_min_eigval}) and (a.s.) satisfy the coherence requirement (\ref{assu:Phi_Q_coherence}) with some $\rho$ and $\kappa$, as well.}

\section{The Sign-Perturbed Sums Algorithm}\label{sec:SPS_overview}
\vspace{-1mm}
In this section we give an overview of the SPS algorithm. {\newtext The core idea behind SPS is to construct the confidence region based on rankings of some sign-perturbed sums, which behave ``similarly'' to the unperturbed sum, when $\theta = \theta^*$, but behave ``differently'' if $\theta$ is ``farther away'' from $\theta^*$. For more detailed intuitions
and for the proofs of the theorems the reader is referred to \cite{Csaji2015} and \cite{Weyer2017}.}
\begin{algorithm}[t]
	\caption{Pseudocode: SPS-Initialization}
     \label{alg:sps_init}
    \begin{flushleft}
    \vspace{1mm}
    {\newtext \textbf{Input:} confidence probability $p$\\
    \textbf{Output:}  auxiliary variables $m,q,\pi,\bar{R}^{-1/2}_n,\{\alpha_{i,t}\}$\\
    \vspace{1mm}
    \textbf{Global variables:} regressors $\{\varphi_t\}$, outputs $\{Y_t\}$}
    \end{flushleft}
    \vspace{1mm}
    \hrule
    \vspace{1mm}
	\begin{algorithmic}[1]
        \STATE Set integers $m > q >0$ such that $p = 1 - q/m$.\\[1mm]
		\STATE Calculate the outer product $\bar{R}_n$ and find the inverse of its principal square root, i.e., the p.s.d.\ matrix
        \vspace{-2mm}
		\begin{equation*}
            \bar{R}_n^{-1/2}\,\bar{R}_n^{-1/2} =\, \bar{R}^{-1}_n.
        \end{equation*}
        \vspace{-6mm}
		\STATE Generate $n\cdot(m-1)$ i.i.d. random signs $\{\alpha_{i,t}\}$ for $i \in \{1,\dots,m-1\}$, $t \in \{1,\dots,n\}$, with:
            \vspace{-2mm}
            \begin{align*}
                \BP(\alpha_{i,t} = 1) \,=\, \BP(\alpha_{i,t} = -1)\, =\, 1/2.
            \end{align*}
            \vspace{-5mm}
		\STATE Generate (uniformly) a random permutation $\pi$ of the set $\{0,\dots, m - 1\}$, where each of the $m!$ possible permutations has probability $1/(m!)$ to be selected.\\[1mm]
	\end{algorithmic}
\end{algorithm}
The SPS algorithm consists of two parts, an initialization phase and an indicator function. In the initialization part the algorithm calculates the main parameters and generates the random signs needed for the construction of the confidence region. The indicator function evaluates whether a given parameter $\theta$ is included in the confidence region{\newtext, i.e., it can be seen as a statistical test for the null hypothesis $\theta^* = \theta$, and then the confidence set is the acceptance region of this test.} 
The initialization algorithm is given in Algorithm \ref{alg:sps_init}, the indicator function is presented in Algorithm \ref{alg:sps_indicator}. Using this construction, the $p$-level SPS confidence region can be defined as follows
\vspace{0.5mm}
\begin{equation}
        \confreg[p,n]\, \defeq\, \{\,\theta \in \BR^d\text{ : SPS-Indicator}(\theta) = 1\hspace{0.3mm}\}.
        \vspace{-3mm}
\end{equation}

\begin{algorithm}[t]
	\caption{Pseudocode: SPS-Indicator}
    \label{alg:sps_indicator}
    \vspace{1mm}
    {\newtext \textbf{Input:} arbitrary parameter vector $\theta$\vspace{0.75mm}\\
    \textbf{Output:} binary decision variable $\beta$\\
    \textbf{Global variables:} $\{\varphi_t\},\{Y_t\},m,q,\pi,\bar{R}^{-1/2}_n,\{\alpha_{i,t}\}$}
    \vspace{1mm}
    \hrule
    \vspace{1mm}
	\begin{algorithmic}[1]
		\STATE Compute the prediction errors for $\theta:$ for $t \in [\hspace{0.3mm}n\hspace{0.2mm}]$ let
        \vspace{-2.5mm}
        \begin{align*}
            \varepsilon_t(\theta)\, \defeq\, Y_t - \varphi_t\tr \theta.
        \end{align*}
        \vspace{-5.5mm}
		\STATE Evaluate for $i \in [\hspace{0.3mm}m-1\hspace{0.2mm}]$ the following functions:
        \vspace{-3mm}
        \begin{align*}
             &S_0(\theta)\, \defeq\, \bar{R}_n^{-\frac{1}{2}}\frac{1}{n}\sum_{t=1}^{n}\varphi_t\varepsilon_t(\theta),\\
             &S_i(\theta)\, \defeq\, \bar{R}_n^{-\frac{1}{2}}\frac{1}{n}\sum_{t=1}^{n}\alpha_{i,t}\varphi_t\varepsilon_t(\theta).
        \end{align*}
        \vspace{-3mm}
		\STATE Compute the rank 
        of $\norm{S_0(\theta)}^2$ among $\{\norm{S_i(\theta)}^2\}:$
        \vspace{-2mm}
            \begin{equation*}
                \mathcal{R}(\theta)\, \defeq\, \left[\,1+\sum_{i=1}^{m-1}\mathbb{I}\left(\norm{S_0(\theta)}^2 \succ_{\pi} \norm{S_i(\theta)}^2\right)\,\right]\!,
                \vspace{-1.5mm}
            \end{equation*}
        where ``$\succ_{\pi}$'' is ``$>$'' with random tie-breaking, i.e., 
        $\norm{S_k(\theta)}^2 \succ_{\pi} \norm{S_j(\theta)}^2$ if and only if $(\norm{S_k(\theta)}^2 > \norm{S_j(\theta)}^2) \lor
        (\norm{S_k(\theta)}^2 = \norm{S_j(\theta)}^2 \,\land\, \pi(k) > \pi(j))$.           
        \vspace{2mm}
        \STATE Set $\beta \defeq 1$, if\, $\mathcal{R}(\theta) \leq m - q$, otherwise set\, $\beta \defeq 0$.
	\end{algorithmic}
\end{algorithm}

It was shown in \cite{Csaji2015} that the confidence region $\confreg[p,n]$ contains the {\newtext true} parameter $\theta^*$ with exact probability $p$, under milder assumptions on the noise and regressors {\newtext than} \ref{assu:noise} and \ref{assu:R_nonsigular}, therefore the following theorem holds:\\
\begin{theorem}\label{thm:exact_confidence}
    Assuming the noise sequence $\{W_t\}$ 
    {\roundtwo contains}
    independent random variables that are symmetric about zero and $\bar{R}_n$ is nonsingular, the confidence probability of the 
    confidence region $\confreg[p,n] $ is exactly $p$,
    \vspace{-1mm}
    \begin{equation}
            \BP(\theta^* \in \confreg[p,n]) \,=\, 1-\frac{q}{m} \,=\, p.
    \end{equation}
\end{theorem}
\vspace{-5mm}
In \cite{Weyer2017} it has been rigorously proven that the confidence regions are also strongly consistent, which requires some further mild assumptions that we do not detail here.

\section{Sample Complexity of SPS}\label{sec:main_result}
\vspace{-1mm}
In this section we give high probability upper bounds for the diameters of the confidence regions generated by the SPS-Indicator algorithm.
An important property of SPS confidence sets is how they shrink as the sample size increases, which is formalized by our main theorem.\\

\begin{theorem}\label{thm:sample_complex_indicator}
    Assuming \ref{assu:noise}, \ref{assu:R_nonsigular}, \ref{assu:R_min_eigval} and \ref{assu:Phi_Q_coherence}, the following concentration inequality holds for the sizes of SPS confidence regions.
    For all\, {\newtext $\delta > 0$ and\, $n \geq \lceil g^{1/\rho}(\frac{\delta}{m-q}) \rceil$} with probability at least\, $1-\delta,$ we have
    \vspace{-1mm}
    \begin{equation}
        \!\!\!\sup_{\theta_1,\theta_2 \in \confreg[p,n]}\!\!\!\!\|\,\theta_1-\theta_2\,\| \,\leq\,
        \dfrac{4\,f\!\left(\frac{\delta}{m-q}\right)}{\left(n^{1-\rho}\lambda_0\!\left(n^{\rho}-g\!\left(\frac{\delta}{m-q}\right)\right)\right)^{\frac{1}{2}}},
    \end{equation}
    where
    {\newtext %
    \begin{align}\label{equ:f_and_g_not_appendix}
            &f(\delta) \,\defeq\, \notag
        \begin{cases}
            \sigma\hspace{0.3mm}(\hspace{0.3mm}8\hspace{0.5mm}d\ln^{\frac{1}{2}}(\tfrac{4}{\delta})+d)^{\frac{1}{2}} & {\newtext 4\e^{-(nd\lambda_0)^2} \leq \delta \leq 2},\\[3mm]
            \sigma\hspace{-0.3mm}\left(8\ln(\tfrac{4}{\delta})+d\right)^{\frac{1}{2}} & {\newtext 0 < \delta < 4\e^{-(nd\lambda_0)^2}} \notag\\
        \end{cases}\\[3mm]
        &g(\delta) \,\defeq\, \ln\left(\tfrac{4d}{\delta}\right)2\hspace{0.3mm}\kappa\hspace{0.3mm} d^2.
    \end{align}}
\end{theorem}
\vspace{-8mm}
In the theorem above $\sup_{\theta_1,\theta_2 \in \confreg[p,n]}\|\theta_1-\theta_2\|$ represents the {\em diameter} of the confidence region. 
The values of $f(\delta)$ and $g(\delta)$ are independent of $n$, hence they can be treated as constants,
{\newtext and $\sigma$ is the variance proxy of the noise (\ref{assu:noise}), hence it is also constant.} 
As a consequence, the decrease rate of SPS 
regions mainly depends on $\rho$ and $n$. It may be surprising that irrespectively of the coherence parameters $\kappa$ and $\rho$, the SPS confidence regions shrink at the {\em optimal} rate, shown by the following corollary.\\

\begin{corollary}\label{cor:asmyp_rate}
    Under the assumptions of Theorem \ref{thm:sample_complex_indicator}, the sizes of the confidence regions generated by the SPS-Indicator algorithm shrink at the rate of \,$\mathcal{O}(1/\sqrt{n})$.
\end{corollary}
\vspace{-3mm}
\begin{proof}
The upper bound in Theorem \ref{thm:sample_complex_indicator} can be rewritten
\begin{align}
    &\dfrac{4f\left(\frac{\delta}{m-q}\right)}{\left(n^{1-\rho}\lambda_0\left(n^{\rho}-g\left(\frac{\delta}{m-q}\right)\right)\right)^{\frac{1}{2}}} \notag\\[3mm]
    &=\; \dfrac{4f\left(\frac{\delta}{m-q}\right)n^{\frac{\rho-1}{2}}}{\left(\lambda_0n^{\rho}-\lambda_0g\left(\frac{\delta}{m-q}\right)\right)^{\frac{1}{2}}},
\end{align}
thus the decrease rate is $\mathcal{O}(n^{\frac{\rho-1}{2}} / n^{\frac{\rho}{2}}) = \mathcal{O}(1/\sqrt{n})$.
\end{proof}
\vspace{-4mm}
Corollary \ref{cor:asmyp_rate} 
is in accordance with the asymptotic results of \cite{Weyer2017}, however, our theorem is {\em non-asymptotic}, since the given 
stochastic 
bound holds for finite $n$ and $m$ values.
Moreover, our theorem describes more precisely the dependence of the diameter on the data characteristics.
\vspace{-3mm}
\begin{proof}[\textbf{Proof of Theorem \ref{thm:sample_complex_indicator}}]
In the first step of the proof we rewrite the SPS confidence region as an ellipsoid for the special case of $m=2$ and $q=1$.
We then decompose the longest axis of the ellipsoid into two terms and give concentration inequalities for both of them. The third step is to give a high probability upper bound for the longest axis of the ellipsoid by combining the previously obtained two concentration inequalities. In the final step, we generalize this result to arbitrary $m$ and $q$ choices.
\vspace*{-1mm}

{\em Step i)} Assume that there are ``two sums'' ($m=2$). The reference $S_0(\theta)$ and the sign-perturbed $S_1(\theta)$ sums are
\begin{align}
    S_0(\theta) 
    &= \bar{R}_n^{-\frac{1}{2}}\frac{1}{n}\sum_{t=1}^n \varphi_t\varepsilon_t(\theta)\notag\\
    &=\bar{R}_n^{-\frac{1}{2}}\frac{1}{n}\sum_{t=1}^n \varphi_t\left(\varphi_t\tr  \theta^* + W_t - \varphi_t\tr \theta\right)\notag\\
    &=\bar{R}_n^{-\frac{1}{2}}\frac{1}{n}\sum_{t=1}^n \varphi_t\varphi_t\tr (\theta^* - \theta) + \varphi_tW_t\notag\\
    &= \bar{R}_n^{\frac{1}{2}}\tilde{\theta} + \tfrac{1}{n}\bar{R}_n^{-\frac{1}{2}} \Phi_n\tr  w_n,\\[-7mm]\notag
\end{align}
\begin{align}
    S_1(\theta) &= \bar{R}_n^{-\frac{1}{2}}\frac{1}{n}\sum_{t=1}^n \alpha_t\varphi_t\varepsilon_t(\theta)\notag\\
    &= \bar{R}_n^{-\frac{1}{2}}\frac{1}{n}\sum_{t=1}^n \alpha_t\varphi_t\left(\varphi_t\tr  \theta^* + W_t - \varphi_t\tr \theta\right) \notag\\
    &=\bar{R}_n^{-\frac{1}{2}}\frac{1}{n}\sum_{t=1}^n \alpha_t\varphi_t\varphi_t\tr (\theta^* - \theta) + \alpha_t\varphi_tW_t\notag\\
    &= \bar{R}_n^{-\frac{1}{2}}\bar{Q}_n\tilde{\theta} + \tfrac{1}{n}\bar{R}_n^{-\frac{1}{2}}\Phi_n\tr  D_{\alpha, n}w_n,\\[-7mm]\notag
\end{align}
where
\begin{equation}
\bar{Q}_n \,\defeq\, \frac{1}{n}Q_n = \frac{1}{n}\sum_{t=1}^n \alpha_t\varphi_t\varphi_t\tr  = \frac{1}{n}\Phi_n\tr  D_{\alpha,n}\Phi_n,
\vspace{-3mm}
\end{equation}
\begin{align}\label{D_alpha}
    &\tilde{\theta} \,\defeq\, \theta^* - \theta, \hspace{13mm}
    D_{\alpha, n} \,\defeq\,
    \begin{bmatrix}
        \alpha_{1} & & \\
        & \ddots & \\
        & & \alpha_{n}
    \end{bmatrix}\!.
\end{align}
The Euclidean norm squares of the sums are
\begin{align}
    &\norm{S_0(\theta)}^2 = \\
    &\left(\bar{R}_n^{\frac{1}{2}}\tilde{\theta} + \tfrac{1}{n}\bar{R}_n^{-\frac{1}{2}} \Phi_n\tr  w_n\right)\tr \left(\bar{R}_n^{\frac{1}{2}}\tilde{\theta} + \tfrac{1}{n}\bar{R}_n^{-\frac{1}{2}} \Phi_n\tr  w_n\right)=\notag\\
    &\frac{1}{n}\left(\tilde{\theta}\tr  R_n\tilde{\theta} + 2\tilde{\theta}\tr \Phi_n\tr  w_n + w_n\tr \Phi_n R_n^{-1}\Phi_n\tr  w_n \right)\!,\notag\\[2mm]
    &\norm{S_1(\theta)}^2 = \left(\bar{R}_n^{-\frac{1}{2}}\bar{Q}_n\tilde{\theta} + \tfrac{1}{n}\bar{R}_n^{-\frac{1}{2}} \Phi_n\tr  D_{\alpha, n}w_n\right)\tr \\
    &\cdot\left(\bar{R}_n^{-\frac{1}{2}}\bar{Q}_n\tilde{\theta} + \tfrac{1}{n}\bar{R}_n^{-\frac{1}{2}} \Phi_n\tr  D_{\alpha, n}w_n\right)
    = \frac{1}{n}\left(\tilde{\theta}\tr  Q_nR_n^{-1}Q_n\tilde{\theta} \right.\notag\\ 
    &\left. + 2\tilde{\theta}\tr  Q_nR_n^{-1}\Phi_n\tr  D_{\alpha,n}w_n + w_n\tr  D_{\alpha,n}\Phi_n R_n^{-1}\Phi_n\tr  D_{\alpha,n}w_n \right)\!.\notag
\end{align}
Parameter $\theta$ is included in the confidence region, for the case of $p=0.5$, if and only if $\norm{S_0(\theta)}^2 \prec_{\pi} \norm{S_1(\theta)}^2$, where ``$\prec_{\pi}$'' is ``$<$'' with random tie-breaking \cite{Csaji2015}. Notice that if we change ``$\prec_{\pi}$'' to ``$\leq$'' we may only include more parameters in the confidence region, therefore the size of the region can only increase. Throughout our proof we will analyse this slightly larger $\norm{S_0(\theta)}^2 - \norm{S_1(\theta)}^2 \leq 0$ region. In other words, we will study the set of those {\newtext $\tilde{\theta}$} vectors which satisfy the constraint
\begin{align}\label{equ:abc_inequ}
    \tilde{\theta}\tr\!  A \,\tilde{\theta} + 2\,\tilde{\theta}\tr  b + c\, \leq\, 0,
\end{align}
where, for simplicity, we did not denote the dependence on the sample size $n$, and used the notation
\begin{align}\label{equ:def_A}
    A &\,\defeq\, R_n - Q_nR_n^{-1}Q_n,\\[1.5mm]
    b &\,\defeq\, B\,w_n,\\[1mm]
    c &\,\defeq\, w_n\tr  C\, w_n,\\[-8mm]\notag
\end{align}
and
\begin{align}\label{equ:def_B_and_C}
    B &\,\defeq\, \Phi_n\tr - Q_nR_n^{-1}\Phi_n\tr  D_{\alpha,n}, \\[2mm]
    C &\,\defeq\, \Phi_n R_n^{-1}\Phi_n\tr - D_{\alpha,n}\Phi_n R_n^{-1}\Phi_n\tr  D_{\alpha,n}.
\end{align}
Note that in \cite{Care2022} it was shown that the SPS confidence regions are bounded if and only if both the perturbed and the unperturbed regressors span the whole space. 
This means that assuming completely exciting regressors (\ref{assu:R_nonsigular}) and $m=2$, the SPS region is bounded if and only if matrix $A$, defined in \eqref{equ:def_A}, is positive definite \cite[Theorem 1]{Care2022}.
In case the region is unbounded, matrix $A$ is positive semidefinite and it has at least one zero eigenvalue.
Inequality (\ref{equ:abc_inequ}) can be reformulated as a (possibly degenerate) ellipsoid even when the region is {\newtext unbounded}
\begin{align}
    &\big\|A^{1/2}(\tilde{\theta} + A^{\dagger}b)\big\|^2 \leq \;b\tr  A^{\dagger}b -c,
\end{align}
{\newtext where we used the notation ``$(\cdot)^\dagger$'' for the pseudoinverse.} Note that our analysis focuses on the size of the SPS confidence region, and the translation (bias) term $A^{\dagger}b$ does not affect the size (volume) of the ellipsoid. 
{\newtext By introducing $\tilde{\theta}_c \defeq \tilde{\theta} + A^{\dagger}b$, using the eigendecomposition of $A = V_{\scriptscriptstyle A}\Lambda_{\scriptscriptstyle A}V_{\scriptscriptstyle A}\tr$ and the fact that
\begin{align}
    \big\|V_{\scriptscriptstyle A}\Lambda_{\scriptscriptstyle A}^{1/2}V_{\scriptscriptstyle A}\tr\tilde{\theta}_c\big\|^2 \geq \lambda_\text{min}(A)\|\tilde{\theta}_c\|^2,
\end{align}
we have}
\begin{align}\label{equ:main_inequ}
    &\|\tilde{\theta}_c\|^2 \,\leq\, \frac{b\tr  A^{\dagger}b -c}{\lambda_\text{min}(A)}\, = \, \frac{\frac{1}{n}w_n\tr  M w_n}{\frac{1}{n}\lambda_\text{min}(A)},
\end{align}
where $M \defeq B\tr  A^{\dagger}B -C$. Notice that $w_n\tr  M w_n$ is lower bounded by some norm, thus $M$ is positive semidefinite. In case the SPS region is unbounded, $\lambda_\text{min}(A) = 0$, therefore, in the above upper bound on the size of the region, there is a division by zero. We consider that in this case the size of the region is infinitely large.
{\em Step ii)} Now that we have reformulated the confidence region as a (generalized) ellipsoid and gave a formula for the longest axis, {\newtext the next step is to give} an upper bound for it. 
{\newtext First, we want to find a lower bound for
$\lambda_{\text{min}}(A)$. To this aim, let us rewrite matrix $A$ as}
\begin{align}\label{equ:reformulation_A}
    A = &\;R_n - Q_nR_n^{-1}Q_n = \Phi_{{\scriptscriptstyle R},n}\tr \Phi_{{\scriptscriptstyle R},n} - \notag\\[1mm]
    &\Phi_{{\scriptscriptstyle R},n}\tr (\Phi_{{\scriptscriptstyle Q},n}\tr  D_{\alpha,n}\Phi_{{\scriptscriptstyle Q},n}\Phi_{{\scriptscriptstyle Q},n}\tr  D_{\alpha,n}\Phi_{{\scriptscriptstyle Q},n})\Phi_{{\scriptscriptstyle R},n},
\end{align}
where $\Phi_n = \Phi_{{\scriptscriptstyle Q},n}\Phi_{{\scriptscriptstyle R},n}$ is the thin QR-decomposition of $\Phi_n$, and $\Phi_{{\scriptscriptstyle R},n}$ is nonsingular ensured by \ref{assu:R_nonsigular}.
Let 
\begin{align}\label{equ:K_and_L_def}
    &K \,\defeq\, \Phi_{{\scriptscriptstyle Q},n}\tr  D_{\alpha,n}\Phi_{{\scriptscriptstyle Q},n},
\end{align}
then, matrix $A$ can be further reformulated as
\begin{align}\label{equ:reformulation_A_2}
    &\Phi_{{\scriptscriptstyle R},n}\tr (I -\Phi_{{\scriptscriptstyle Q},n}\tr  D_{\alpha,n}\Phi_{{\scriptscriptstyle Q},n}\Phi_{{\scriptscriptstyle Q},n}\tr  D_{\alpha,n}\Phi_{{\scriptscriptstyle Q},n})\Phi_{{\scriptscriptstyle R},n} = \notag\\[1mm] 
    &\Phi_{{\scriptscriptstyle R},n}\tr (I - K^2)\Phi_{{\scriptscriptstyle R},n}.
\end{align}
Note that $A$ is a positive semidefinite matrix, therefore $I-K^2$ is also positive semidefinite, thus it holds for the eigenvalues of $K$ that $\forall\, i : \abs{\lambda_i(K)} \leq 1$. Also note that from the connection between the eigenvalues of $A$ and the boundedness of the region, detailed previously, it follows that $K$ has an eigenvalue of value 1, if and only if the region is unbounded. From \ref{assu:R_nonsigular} it follows that $K$ is nonsingular, therefore $\forall\, i : 0 < \abs{\lambda_i(K)}$. Then, $\lambda_{\text{min}}(A)=\lambda_{\text{min}}(\Phi_{{\scriptscriptstyle R},n}\tr(I-K^2)\Phi_{{\scriptscriptstyle R},n})$ can be reformulated using the eigendecomposition
\begin{align}
 &\lambda_{\text{min}}(\Phi_{{\scriptscriptstyle R},n}\tr(I-K^2)\Phi_{{\scriptscriptstyle R},n})=\notag\\[1mm]
 &v_{\text{min}}\tr \Phi_{{\scriptscriptstyle R},n}\tr(I-K^2)\Phi_{{\scriptscriptstyle R},n}v_{\text{min}},
\end{align}
where $v_{\text{min}}$ is a unit length eigenvector of $A$ corresponding to the smallest eigenvalue of $A$, $\lambda_{\text{min}}(A)$. 
A lower bound on the above expression can be given as
\begin{align}
&v_{\text{min}}\tr \Phi_{{\scriptscriptstyle R},n}\tr(I-K^2)\Phi_{{\scriptscriptstyle R},n}v_{\text{min}} \geq \notag\\[1mm]
&\lambda_{\text{min}}(I-K^2)\norm{\Phi_{{\scriptscriptstyle R},n}v_{\text{min}}}^2 \geq\notag\\[1mm]
&\lambda_{\text{min}}(I-K^2)\sigma_{\text{min}}^2(\Phi_{{\scriptscriptstyle R},n})\norm{v_{\text{min}}}^2,
\end{align}
where $\sigma_{\text{min}}(\Phi_{{\scriptscriptstyle R},n})$ is the smallest singular value of $\Phi_{{\scriptscriptstyle R},n}$.
Since $\norm{v_{\text{min}}}^2 = 1$ and $R_n = \Phi_n\tr \Phi_n = \Phi_{{\scriptscriptstyle R},n}\tr \Phi_{{\scriptscriptstyle R},n}$ we get
\begin{align}
\lambda_{\text{min}}(A) &\geq\lambda_{\text{min}}(I-K^2)\sigma_{\text{min}}^2(\Phi_{{\scriptscriptstyle R},n})\norm{v_{\text{min}}}^2 \notag\\[1mm]
&= \lambda_{\text{min}}(I-K^2)\lambda_{\text{min}}(R_n).
\end{align}
Using the above lower bound on $\lambda_{\text{min}}(A)$, assumption \ref{assu:R_min_eigval} and the fact that $\lambda_{\text{min}}(I-K^2) = 1-\lambda_{\text{max}}(K^2)$, the following upper bound can be given for $\|\tilde{\theta}_c\|^2$
\begin{align}
    \| \tilde{\theta}_c\|^2 &\leq \frac{\frac{1}{n}w_n\tr  M w_n}{\frac{1}{n}\lambda_\text{min}(A)} \leq \frac{\frac{1}{n}w_n\tr  M w_n}{\frac{1}{n}(1-\lambda_{\text{max}}(K^2))\lambda_{\text{min}}(R_n)}  \notag\\[1mm]
    &\leq \frac{\frac{1}{n}w_n\tr  M w_n}{\lambda_0(1-\lambda_{\text{max}}(K^2))}.
\end{align}
Now, we will investigate the non-asymptotic behavior of $(w_n\tr  M w_n)/(n\lambda_0(1-\lambda_{\text{max}}(K^2)))$ as the sample size increases. First, we decompose the previous fraction as
\begin{align}\label{equ:indicator_ub_fraction_form}
    \frac{w_n\tr  M w_n}{n\lambda_0(1-\lambda_{\text{max}}(K^2))} = \frac{w_n\tr  M w_n}{n\lambda_0}\frac{1}{1-\lambda_{\text{max}}(K^2)},
\end{align}
and study the two terms separately. We claim that $M$ is an orthogonal projection matrix with rank at most $d:$
\begin{lemma}\label{lemma:M_proj}
    Assuming \ref{assu:R_nonsigular}, the matrix $M = B\tr  A^{\dagger}B -C$ 
    is an orthogonal projection matrix with rank($M$) $\leq d$.
\end{lemma}
\vspace{-4mm}
The proof of Lemma \ref{lemma:M_proj} is presented in Appendix \ref{sec:proof_M_proj}.
Then, for the first term, namely $(w_n\tr  M w_n)/(n\lambda_0)$, the following concentration inequality can be stated.
\begin{lemma}\label{lemma:numerator_indicator}
    Assuming \ref{assu:noise}, \ref{assu:R_nonsigular}
    and that $M$ is an orthogonal projection matrix with rank($M$) $\leq d,$ the following concentration inequality holds for the random variable $X = w_n\tr  M w_n,$ for every\, $\varepsilon \geq 0:$
    \begin{align}
        &\BP\left(\frac{\vert X - \mathbb{E}X\vert}{n\lambda_0} \geq \varepsilon\right) \leq \notag\\[2mm]
        &\begin{cases}
            2\exp(-\frac{\varepsilon^2n^2\lambda_0^2}{64d^2\sigma^4}) & 0 \leq \varepsilon \leq 8\sigma^2 d^2\\[2mm]
            2\exp(-\frac{\varepsilon n\lambda_0}{8\sigma^2}) & \varepsilon > 8\sigma^2{d^2}.\\[0.5mm]
        \end{cases}
    \end{align}
\end{lemma}
\vspace{-4mm}
The proof of Lemma \ref{lemma:numerator_indicator} can be found in Appendix \ref{sec:proof_num_indicator}. Next we give a concentration inequality for the second term.
Since $\abs{\lambda_i(K)}\in(0,1]$, we investigate the probability for every $\varepsilon > 1:$
\begin{align}\label{equ:indicator_sec_term_prob}
    &\BP\left(\frac{1}{1-\lambda_{\text{max}}(K^2)}\geq \varepsilon\right) = \BP\left(\frac{1}{\varepsilon}\geq 1-\lambda_{\text{max}}(K^2)\right) = \notag
    \\[2mm]
    &\BP\left(\max_i\abs{\lambda_i(K)}\geq \sqrt{1-\frac{1}{\varepsilon}}\right).
\end{align}
Recall that the confidence region is unbounded if and only if $\lambda_{\text{max}}(K) = 1$, in that case, as we mentioned before, we consider ``$1/0$''$ = \infty$. The following lemma gives a concentration inequality for $\max_i\abs{\lambda_i(K)}$.
\begin{lemma}\label{lemma:denominator_indicator}
    Assuming \ref{assu:R_nonsigular} and \ref{assu:Phi_Q_coherence}, the following concentration inequality holds for every $0 < \varepsilon_0 \leq 1:$
    \begin{equation}
        \BP\left(\max_i \abs{\lambda_i(K)} \geq \varepsilon_0\right) \leq\, 2\,d\exp\!\left(-\frac{n^{\rho}\varepsilon_0^2}{2\kappa d^2}\right).
    \end{equation}
\end{lemma}
\vspace{-4mm}
The formal proof of Lemma \ref{lemma:denominator_indicator} is given in Appendix \ref{sec:proof_den_indicator}. 

{\em Step iii)} Using Lemma \ref{lemma:denominator_indicator} and the reformulation of \eqref{equ:indicator_sec_term_prob}, the following probability upper bound can be given for the second term $1/(1-\lambda_{\text{max}}(K^2))$, for every $\varepsilon > 1:$
\begin{equation}\label{equ:indicator_sec_term_inequ}
    \hspace*{-2mm}\BP\left(\frac{1}{1-\lambda_{\text{max}}(K^2)}\geq \varepsilon\right) \leq
    2d\exp\left(-\frac{n^{\rho}\left(1-\frac{1}{\varepsilon}\right)}{2\kappa d^2}\right)\!.\!
\end{equation}
The next lemma gives a concentration inequality for the upper bound of the longest axis of the ellipsoid, $(w_n\tr  M w_n)/(n\lambda_0(1-\lambda_{\text{max}}(K^2))$, by using the results of Lemma \ref{lemma:numerator_indicator} and \eqref{equ:indicator_sec_term_inequ}. Note that this lemma gives a concentration inequality for the size of the confidence region generated by the SPS algorithm for $m=2$ and $q=1$.
\vspace{1mm}
\begin{lemma}\label{lemma:sample_complex_indicator_m2_q1}
    Assuming \ref{assu:noise}-\ref{assu:Phi_Q_coherence}, the following concentration inequality holds for the sizes of the confidence regions generated by the SPS-Indicator algorithm for $m=2$ and $q=1$, with probability at least $1-\delta:$
    \begin{align}\label{equ:m2_sample_complex}
        \sup_{\theta_1,\theta_2 \in \confreg[0.5,n]}\!\!\!\!\!\|\theta_1-\theta_2\| \,\leq\,
        \dfrac{2f(\delta)}{\left(n^{1-\rho}\lambda_0\left(n^{\rho}-g(\delta)\right)\right)^{1/2}}.
    \end{align}
\end{lemma}
\vspace{-1mm}
The proof of Lemma \ref{lemma:sample_complex_indicator_m2_q1} and the definitions of functions $f$ and $g$ are given in Appendix \ref{sec:proof_sample_complex_indicator_m2_q1}.
\begin{remark}
It is important to note that the stochastic upper bound of Lemma \ref{lemma:denominator_indicator} also covers the case of unbounded confidence regions, since the unbounded case can arise if and only if\, $\lambda_{\max}(K)=1$. Hence, the unbounded case is also covered by Lemma \ref{lemma:sample_complex_indicator_m2_q1}, as it builds on Lemma \ref{lemma:denominator_indicator}. 
\end{remark}
\vspace{-4mm}
{\em Step iv)} As the final step, we consider the general case: we allow arbitrary $m > q > 0$ (integer) choices. Our aim will be to provide an upper bound for the probability of the ``bad'' event that the size of the constructed region is above $2f(\delta)/(n^{1-\rho}\lambda_0\left(n^{\rho}-g(\delta)\right))^{1/2}$. 
First, we can construct ``good'' events for each $i \in [m-1]$, i.e., the event that the $0.5$ probability region defined by $S_0(\theta)$ and $S_i(\theta)$ is upper bounded according to \eqref{equ:m2_sample_complex}. This event is\vspace{-1mm}
\begin{align}\label{equ:arbitrary_mq_start}
    \CG_i\, \defeq \,\biggl\{\,\omega \in \Omega : &\sup_{{\theta}_{i,1},{\theta}_{i,2} \in \confreg[0.5]^i(\omega)}\|{\theta}_{i,1}-{\theta}_{i,2}\| \leq \notag\\[1mm]
    &\dfrac{2f(\delta)}{\left(n^{1-\rho}\lambda_0\left(n^{\rho}-g(\delta)\right)\right)^{1/2}}\,\biggr\},
\end{align}
where ${\theta}_{i,1}$ and ${\theta}_{i,2}$ are points included in the $0.5$-level confidence set $\confreg[0.5]^i \doteq \{\, \theta : 
\|S_0(\theta)\|^2 \leq \|S_i(\theta)\|^2\,\},$
and $\Omega$ is the sample space of the underlying probability space $(\Omega, \mathcal{F}, \BP)$. We already provided a lower bound for $\BP(\CG_i)$, see \eqref{equ:m2_sample_complex}, which is valid for all $i \in [m-1]$; but $\CG_1, \dots, \CG_{m-1}$ are not independent. By using the construction of SPS confidence regions, see \cite{Csaji2015}, the good event, given integers $0 < q < m$, can be rewritten as
\vspace{-0.5mm}
\begin{equation}\label{equ:good_event_construct}
\CG\; \defeq\hspace{-1mm} \bigcup_{\substack{I \subseteq [m-1],\\ |I| \geq m-q}}\hspace{1mm} \bigcap_{i \in I} \;\CG_i.
\vspace{-0.5mm}
\end{equation}
Note that in our analysis we investigate the region 
$\confreg[0.5]^i$,
which is slightly larger than the SPS region.
By the construction of the SPS confidence sets, we have $\|S_0(\hat{\theta}_n)\|^2 = 0$, 
where $\hat{\theta}_n \defeq R_n^{-1}\Phi_n\tr y_n$ is the least squares estimate (LSE).
Therefore, this (enlarged) set is not empty, since the LSE is always included in the region. Then, \eqref{equ:good_event_construct} means that there exist at least $m-q$ (perturbed) paraboloids, $\{\|S_i(\theta)\|^2\}_{i \neq 0}$, such that any point in all of their intersections with the reference $\|S_0(\theta)\|^2$ are closer to $\hat{\theta}_n$ than $2f(\delta)/(n^{1-\rho}\lambda_0\left(n^{\rho}-g(\delta)\right))^{1/2}$, therefore any two points in $\CG$ are closer than $4f(\delta)/(n^{1-\rho}\lambda_0\left(n^{\rho}-g(\delta)\right))^{1/2}$.

Then, by using De Morgan's laws, the  ``bad'' event is
\begin{equation}
\CB\, \defeq \; \Omega \setminus \CG \,=\hspace{-1mm} \bigcap_{\substack{I \subseteq [m-1],\\ |I| \geq m-q}}\hspace{1mm} \bigcup_{i \in I} \;\CB_i \,=\hspace{-1mm} \bigcap_{\substack{I \subseteq [m-1],\\ |I| = m-q}}\hspace{1mm} \bigcup_{i \in I} \;\CB_i,
\end{equation}
where $\CB_i = \Omega \setminus \CG_i$, for $i\in[m-1]$.
The probability of this ``bad'' event can be bounded by
\vspace{-1mm}
\begin{equation}
\BP(\CB)\, \leq \min_{\substack{I \subseteq [m-1],\\ |I| = m-q}}\hspace{0mm} \BP\!\left[\,\bigcup_{i \in I}\,\CB_i \right] \, \leq\, (m-q)\cdot \BP(\CB_1),
\end{equation}
where we used that $\BP(A \cap B) \leq \min\{\BP(A),\, \BP(B)\}$ and $\BP(A \cup B) \leq \BP(A) + \BP(B)$. An upper bound on $\BP(\CB_1)$ can be given by using the lower bound on $\BP(\CG_1)$ from \eqref{equ:m2_sample_complex} as
\begin{equation}
    1-\BP(\CB_1) = \BP(\CG_1) \geq 1 - \delta\quad \Longrightarrow\quad \BP(\CB_1) \leq \delta,    
\end{equation}
{\newtext therefore $\BP(\CB) \leq (m-q)\delta$.
By introducing $\delta' = (m-q)\delta$ it can be written that
$
    1-\delta' \leq 1-\BP(\CB) = \BP(\CG)
$, hence}
\vspace{-1mm}
\begin{align}\label{equ:sample_complex_indicator}
    \sup_{\theta_1,\theta_2 \in \confreg[p,n]}\|\theta_1-\theta_2\| &\leq
    \dfrac{4f(\delta)}{\left(n^{1-\rho}\lambda_0\left(n^{\rho}-g(\delta)\right)\right)^{1/2}}\\
   &=\dfrac{4f(\frac{\delta'}{m-q})}{\left(n^{1-\rho}\lambda_0\left(n^{\rho}-g(\frac{\delta'}{m-q})\right)\right)^{1/2}},\notag
\end{align}
{\newtext with probability at least $1-\delta'$.}
\end{proof}
\section{Simulation Experiments}\label{sec:experiments}
\vspace{-1mm}
In this section we compare our theoretical bounds on the sizes of the confidence regions with the sizes of the regions given by simulated trajectories. We consider a 2-dimensional system, given in the form of \eqref{equ:system} with $\theta^* = [5, 5]\tr$,  $W_t \sim \text{Unif}(-1,1)$ and $\varphi_{t,i} \sim \text{Unif}(1,2)$. Throughout our experiments we consider $0.5$-level confidence regions, that is $m = 2$ and $q=1$, a sample size of $n = 2000$ and $s = 100$ independently simulated trajectories. We set $\delta = 0.1$, $\rho=1$ and from the noise setting it follows that $\sigma^2 = 1/3$ is the optimal variance proxy.

\begin{figure}[t!]
\vspace{1mm}
\begin{center}
\includegraphics[width=\columnwidth]{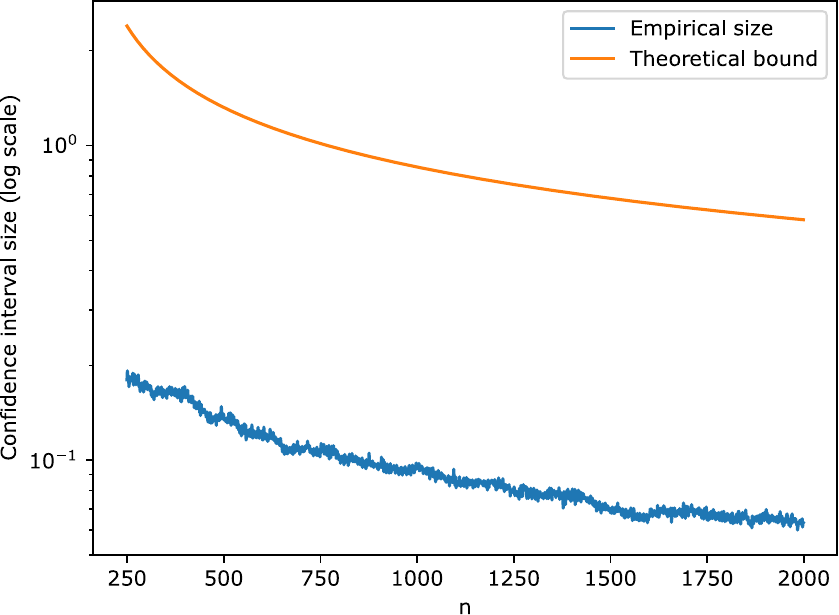}    %
\caption{Comparison of the empirical sizes and the theoretical upper bound on the sizes of the SPS-Indicator region for $m=2$, $\delta=0.1$, $t_0=250$, $n=2000$ and $s =100$.}
\label{fig:t025}
\vspace{3mm}
\end{center}                                 
\end{figure}

The remainder constants that appear in our theoretical bound are computed from the data, that is, $\lambda_0$ is set as the smallest empirical eigenvalue of $\bar{R}_{t}$ and $\kappa$ as the largest empirical value of $\frac{t}{d} \max_{1\leq i \leq t}\norm{\Phi_{Q,t}\tr e_i}^2$ over all $t_0\leq t \leq n$ and simulated trajectories.

During the simulations, $100$ random points are sampled from the SPS-Indicator region. Then, the empirical diameter corresponding to $\sup_{\theta_1,\theta_2 \in \confreg[p,n]}\|\theta_1-\theta_2\|$ is the maximal Euclidean distance of all the sampled points.

The difference between the empirical size with confidence level $1-\delta$ and the theoretical size are shown in Fig. \ref{fig:t025}. The theoretical bound is computed according to the result of Theorem \ref{thm:sample_complex_indicator}. For the empirical size, quantiles were used for each iteration, i.e., the smallest number for which at least the specified portion of simulation realizations are below that number.

The results are indicative of the phenomenon that our theoretical rate captures well the empirical decrease rate of the diameter of SPS confidence regions. On the other hand, the bounds appear conservative. This, however, comes from the fact that these bounds can be computed {\em a priori}, as they are based on concentration inequalities, hence they cannot be as efficient as SPS, which is a {\em data-driven} method to build {\em exact} confidence regions.

\section{Conclusion}\label{sec:conclusion}
\vspace{-1mm}
We have analyzed the {\em sample complexity} of the Sign-Perturbed Sums (SPS) finite-sample, distribution-free system identification method. We have focused on linear regression problems and assumed that the measurement noises are independent, symmetric and subgaussian, as well as that the regressors are suitably exciting. 

We have proven a {\em non-asymptotic} concentration bound which shows how the sizes of SPS confidence regions depend on the data characteristics and the significance level. As a corollary, we have shown that the diameters of SPS regions shrink at the same optimal rate as the confidence ellipsoids based on the asymptotic theory. 

Future research directions include extending the results to the ellipsoidal outer-approximation of SPS, and to dynamical systems, even in closed-loop settings.

\bibliographystyle{elsarticle-num}        %
\bibliography{autosam}           %

\appendix
\section{Proof of Lemma \ref{lemma:M_proj}}\label{sec:proof_M_proj}
\vspace{-1mm}
{\newtext Since $M = B\tr  A^{\dagger}B -C$ and both $A$ and $C$ are symmetric by definition, it is clear that $M = M\tr $.} Note that if $M=0$ the Lemma trivially holds, therefore in the following analysis we assume that $M$ has at least one non-zero eigenvalue. We reformulate the matrices $B = \Phi_n\tr - Q_nR_n^{-1}\Phi_n\tr  D_{\alpha,n}$ and $C = \Phi_n R_n^{-1}\Phi_n\tr - D_{\alpha,n}\Phi_n R_n^{-1}\Phi_n\tr  D_{\alpha,n}$ from \eqref{equ:def_B_and_C} using the same ideas as in the reformulation of $A$ in \eqref{equ:reformulation_A}-\eqref{equ:reformulation_A_2}
{\newtext (specifically, using the thin QR decomposition $\Phi_n = \Phi_{{\scriptscriptstyle Q},n}\Phi_{{\scriptscriptstyle R},n}$ and the definition of $K = \Phi_{{\scriptscriptstyle Q},n}\tr  D_{\alpha,n}\Phi_{{\scriptscriptstyle Q},n}$) as}
\vspace{-1mm}
\begin{align}
    &A = \Phi_{{\scriptscriptstyle R},n}\tr (I-K^2) \Phi_{{\scriptscriptstyle R},n},\notag\\
    &B = \Phi_{{\scriptscriptstyle R},n}\tr\Phi_{{\scriptscriptstyle Q},n}\tr- \Phi_{{\scriptscriptstyle R},n}\tr K \Phi_{{\scriptscriptstyle Q},n}\tr D_{\alpha,n},\notag\\
    &C = \Phi_{{\scriptscriptstyle Q},n}\Phi_{{\scriptscriptstyle Q},n}\tr - D_{\alpha,n}\Phi_{{\scriptscriptstyle Q},n}\Phi_{{\scriptscriptstyle Q},n}\tr D_{\alpha,n}.
\end{align}
Then, by using that $\Phi_{{\scriptscriptstyle R},n}$ is nonsingular (\ref{assu:R_nonsigular}), $M = B\tr  A^{\dagger}B -C$ can be rewritten as
\begin{align}
    M &= B\tr  A^{\dagger}B -C \notag\\
    &=\left(\Phi_{{\scriptscriptstyle Q},n}\Phi_{{\scriptscriptstyle R},n}- D_{\alpha,n}\Phi_{{\scriptscriptstyle Q},n}K\Phi_{{\scriptscriptstyle R},n}\right)\Phi_{{\scriptscriptstyle R},n}^{-1}(I-K^2)^{\dagger}\notag\\
    &\quad\; \cdot\Phi_{{\scriptscriptstyle R},n}\ntr\big(\Phi_{{\scriptscriptstyle R},n}\tr\Phi_{{\scriptscriptstyle Q},n}\tr- \Phi_{{\scriptscriptstyle R},n}\tr K \Phi_{{\scriptscriptstyle Q},n}\tr D_{\alpha,n}\big) \notag\\
    &\quad\; -\Phi_{{\scriptscriptstyle Q},n}\Phi_{{\scriptscriptstyle Q},n}\tr + D_{\alpha,n}\Phi_{{\scriptscriptstyle Q},n}\Phi_{{\scriptscriptstyle Q},n}\tr D_{\alpha,n}\\
    &=\Phi_{{\scriptscriptstyle Q},n} \left((I-K^2)^{\dagger} - I\right)\Phi_{{\scriptscriptstyle Q},n}\tr - \Phi_{{\scriptscriptstyle Q},n} (I-K^2)^{\dagger}\notag\\
    &\quad\; \cdot K\Phi_{{\scriptscriptstyle Q},n}\tr D_{\alpha,n}-D_{\alpha,n}\Phi_{{\scriptscriptstyle Q},n}K(I-K^2)^{\dagger}\Phi_{{\scriptscriptstyle Q},n}\tr\notag\\
    &\quad\; + D_{\alpha,n}\Phi_{{\scriptscriptstyle Q},n}\left(K(I-K^2)^{\dagger}K + I \right)\Phi_{{\scriptscriptstyle Q},n}\tr D_{\alpha,n}.\notag
\end{align}
Now, we rewrite $(I-K^2)^{\dagger} - I$ and $K(I-K^2)^{\dagger}K + I$ from the reformulation of $M$. Recall from the proof of Theorem \ref{thm:sample_complex_indicator} that $0 < \abs{\lambda_i(K)} \leq 1 \quad \forall\, i$, consequently $\text{rank}(K) = \text{rank}(\Phi_{{\scriptscriptstyle Q},n}) = d$ (\ref{assu:R_nonsigular}). By denoting the eigendecomposition of $K$ by $K = V_{\scriptscriptstyle K}\Lambda_{\scriptscriptstyle K}V_{\scriptscriptstyle K}\tr$, we have
\vspace{-1mm}
\begin{align}
    &(I-K^2)^{\dagger}-I = V_{\scriptscriptstyle K}\left(\left(I-\Lambda_{\scriptscriptstyle K}^2\right)^{\dagger}-I\right)V_{\scriptscriptstyle K}\tr = \notag\\
    &= V_{\scriptscriptstyle K} \text{diag}\Big(\frac{1}{1-\lambda_1^2(K)}-1, \dots, \frac{1}{1-\lambda_{d'}^2(K)}-1,\notag\\
    &\quad\; -1,\dots,-1\Big)V_{\scriptscriptstyle K}\tr \notag\\
    &= V_{\scriptscriptstyle K} \text{diag}\Big(\frac{\lambda_1^2(K)}{1-\lambda_1^2(K)}, \dots, \frac{\lambda_{d'}^2(K)}{1-\lambda_{d'}^2(K)},\notag\\
    &\quad\; -1,\dots,-1\Big)V_{\scriptscriptstyle K}\tr\notag\\
    &= V_{\scriptscriptstyle K}\left(\left(I-\Lambda_{\scriptscriptstyle K}^2\right)^{\dagger}\Lambda_{\scriptscriptstyle K}^2 - D_{1}\right)V_{\scriptscriptstyle K}\tr\notag\\
    &= (I-K^2)^{\dagger}K^2 - V_{\scriptscriptstyle K}D_{1}V_{\scriptscriptstyle K}\tr,
\end{align}
and
\begin{align}
    &K(I-K^2)^{\dagger}K+I = V_{\scriptscriptstyle K}\left(\left(I-\Lambda_{\scriptscriptstyle K}^2\right)^{\dagger}\Lambda_{\scriptscriptstyle K}^2+I\right)V_{\scriptscriptstyle K}\tr \notag\\
    &=\; V_{\scriptscriptstyle K} \text{diag}\Big(\frac{\lambda_1^2(K)}{1-\lambda_1^2(K)}+1, \dots, \frac{\lambda_{d'}^2(K)}{1-\lambda_{d'}^2(K)}+1,\notag\\
    &\quad\;\;\,1,\dots,1\Big)V_{\scriptscriptstyle K}\tr \notag\\
    &= V_{\scriptscriptstyle K} \text{diag}\left(\frac{1}{1-\lambda_1^2(K)}, \dots, \frac{1}{1-\lambda_{d'}^2(K)},1,\dots,1\right)V_{\scriptscriptstyle K}\tr \notag\\
    &=V_{\scriptscriptstyle K}\left(\left(I-\Lambda_{\scriptscriptstyle K}^2\right)^{\dagger} + D_{1}\right)V_{\scriptscriptstyle K}\tr \notag\\[1mm]
    & = (I-K^2)^{\dagger} + V_{\scriptscriptstyle K}D_{1}V_{\scriptscriptstyle K}\tr,
\end{align}
where $0 < d' \leq d$ and $D_{1} = \text{diag}(0_1,\dots,0_{d'},1,\dots,1)$.
From the above eigendecompositions of $K$ it follows that the matrices {\newtext$(I-K^2)^{\dagger}$} and $K$ commute, thus
\begin{align}
    M &= \Phi_{{\scriptscriptstyle Q},n}\left( (I-K^2)^{\dagger}K^2 - V_{\scriptscriptstyle K}D_{1}V_{\scriptscriptstyle K}\tr \right)\Phi_{{\scriptscriptstyle Q},n}\tr \notag\\
    &\quad\; - \Phi_{{\scriptscriptstyle Q},n} (I-K^2)^{\dagger}K\Phi_{{\scriptscriptstyle Q},n}\tr D_{\alpha,n}\notag\\
    &\quad\; -D_{\alpha,n}\Phi_{{\scriptscriptstyle Q},n}K(I-K^2)^{\dagger}\Phi_{{\scriptscriptstyle Q},n}\tr \notag\\
    &\quad\; + D_{\alpha,n}\Phi_{{\scriptscriptstyle Q},n}\left((I-K^2)^{\dagger} + V_{\scriptscriptstyle K}D_{1}V_{\scriptscriptstyle K}\tr\right)\Phi_{{\scriptscriptstyle Q},n}\tr D_{\alpha,n}\notag\\[1mm]
    &=\left(\Phi_{{\scriptscriptstyle Q},n}K-D_{\alpha,n}\Phi_{{\scriptscriptstyle Q},n}\right)(I-K^2)^{\dagger}\notag\\
    &\quad\; \cdot\left(\Phi_{{\scriptscriptstyle Q},n}K-D_{\alpha,n}\Phi_{{\scriptscriptstyle Q},n}\right)\tr- \Phi_{{\scriptscriptstyle Q},n}V_{\scriptscriptstyle K}D_{1}V_{\scriptscriptstyle K}\tr\Phi_{{\scriptscriptstyle Q},n}\tr\notag\\
    &\quad\; +D_{\alpha,n}\Phi_{{\scriptscriptstyle Q},n}V_{\scriptscriptstyle K}D_{1}V_{\scriptscriptstyle K}\tr\Phi_{{\scriptscriptstyle Q},n}\tr D_{\alpha,n}.
\end{align}
In the following few steps, we show that 
\begin{align}
    &-\Phi_{{\scriptscriptstyle Q},n}V_{\scriptscriptstyle K}D_{1}V_{\scriptscriptstyle K}\tr\Phi_{{\scriptscriptstyle Q},n}\tr + D_{\alpha,n}\Phi_{{\scriptscriptstyle Q},n}V_{\scriptscriptstyle K}D_{1}V_{\scriptscriptstyle K}\tr\Phi_{{\scriptscriptstyle Q},n}\tr D_{\alpha,n}\notag\\[1mm]
    &= 0.
\end{align}
Notice that $\Phi_{{\scriptscriptstyle Q},n}V_{\scriptscriptstyle K}$ is the eigenvector matrix of $\Phi_{{\scriptscriptstyle Q},n}V_{\scriptscriptstyle K}D_{1}V_{\scriptscriptstyle K}\tr\Phi_{{\scriptscriptstyle Q},n}\tr$, since $V_{\scriptscriptstyle K}\tr\Phi_{{\scriptscriptstyle Q},n}\tr\Phi_{{\scriptscriptstyle Q},n}V_{\scriptscriptstyle K} = I$ and $D_{1}$ is diagonal, furthermore $D_{\alpha,n}\Phi_{{\scriptscriptstyle Q},n}V_{\scriptscriptstyle K}$ is the eigenvector matrix of $D_{\alpha,n}\Phi_{{\scriptscriptstyle Q},n}V_{\scriptscriptstyle K}D_{1}V_{\scriptscriptstyle K}\tr\Phi_{{\scriptscriptstyle Q},n}\tr D_{\alpha,n}$ with eigenvalue matrix $D_{1}$ because of the same reasons. It can be shown that $\Phi_{{\scriptscriptstyle Q},n}V_{\scriptscriptstyle K}$ is also an eigenvector matrix for $D_{\alpha,n}\Phi_{{\scriptscriptstyle Q},n}V_{\scriptscriptstyle K}D_{1}V_{\scriptscriptstyle K}\tr\Phi_{{\scriptscriptstyle Q},n}\tr D_{\alpha,n}$, since by multiplying with $V_{\scriptscriptstyle K}\tr\Phi_{{\scriptscriptstyle Q},n}\tr$ from the left and $\Phi_{{\scriptscriptstyle Q},n}V_{\scriptscriptstyle K}$ from the right, using the same ordering of eigenvalues and eigenvectors and $K = V_{\scriptscriptstyle K}\Lambda_{\scriptscriptstyle K}V_{\scriptscriptstyle K}\tr$ we get
\begin{align}
    &V_{\scriptscriptstyle K}\tr\Phi_{{\scriptscriptstyle Q},n}\tr D_{\alpha,n}\Phi_{{\scriptscriptstyle Q},n}V_{\scriptscriptstyle K}D_{1}V_{\scriptscriptstyle K}\tr\Phi_{{\scriptscriptstyle Q},n}\tr D_{\alpha,n}\Phi_{{\scriptscriptstyle Q},n}V_{\scriptscriptstyle K} =\notag\\[1mm]
    &= V_{\scriptscriptstyle K}\tr KV_{\scriptscriptstyle K}D_{1}V_{\scriptscriptstyle K}\tr K V_{\scriptscriptstyle K}= \Lambda_{\scriptscriptstyle K}D_{1}\Lambda_{\scriptscriptstyle K} = D_{1}.
\end{align}
Because the two matrices $D_{\alpha,n}\Phi_{{\scriptscriptstyle Q},n}V_{\scriptscriptstyle K}D_{1}V_{\scriptscriptstyle K}\tr\Phi_{{\scriptscriptstyle Q},n}\tr D_{\alpha,n}$ and $\Phi_{{\scriptscriptstyle Q},n}V_{\scriptscriptstyle K}D_{1}V_{\scriptscriptstyle K}\tr\Phi_{{\scriptscriptstyle Q},n}\tr$ can be written with the same eigenvectors and eigenvalues, it follows that they are equal, therefore their difference is zero. Then, 
\begin{align}\label{equ:M_as_product}
    M =&\left(\Phi_{{\scriptscriptstyle Q},n}K-D_{\alpha,n}\Phi_{{\scriptscriptstyle Q},n}\right)(I-K^2)^{\dagger}\notag\\[1mm]
       &\cdot\left(\Phi_{{\scriptscriptstyle Q},n}K-D_{\alpha,n}\Phi_{{\scriptscriptstyle Q},n}\right)\tr.
\end{align}
Next, we will prove that $M$ is a projection matrix. First, we introduce the notation $M_0 \defeq \Phi_{{\scriptscriptstyle Q},n}K-D_{\alpha,n}\Phi_{{\scriptscriptstyle Q},n}$. 
Then, notice that
\begin{align}
    M_0\tr M_0 &= \left(\Phi_{{\scriptscriptstyle Q},n}K-D_{\alpha,n}\Phi_{{\scriptscriptstyle Q},n}\right)\tr\left(\Phi_{{\scriptscriptstyle Q},n}K-D_{\alpha,n}\Phi_{{\scriptscriptstyle Q},n}\right) \notag\\[2mm]
    &=K^2-K^2-K^2+I = I-K^2.
\end{align}
Writing $M_0$ back to \eqref{equ:M_as_product} we conclude that $M$ is
\begin{align}
    M = M_0(M_0\tr M_0)^{\dagger} M_0\tr,
\end{align}
which is an orthogonal projection 
to the column space of $M_0$, as one can easily verify that $M = M\tr$ and $M^2 = M.$
From the above form of $M$, it also follows that $\text{rank}(M) \leq \text{rank}(M_0) \leq d$, since the rank cannot be increased by matrix products, and $M_0 \in \mathbb{R}^{n \times d}$, $d \leq n$.\qed
\section{Proof of Lemma \ref{lemma:numerator_indicator}}\label{sec:proof_num_indicator}
    \vspace{-1mm}
    Matrix $M$ is symmetric and positive semidefinite,
    hence it can be factorized using the eigendecomposition as
    \begin{align}
        X = w_n\tr  M w_n = w_n\tr  V_{\scriptscriptstyle M} \Lambda_{\scriptscriptstyle M} V_{\scriptscriptstyle M}\tr  w_n.
    \end{align}
    Since $M$ is a projection matrix and rank($M$) $ = d' \leq d$, it follows that $M$ has $d'$ eigenvalues of 1 and $n-d'$ eigenvalues of 0, therefore $\Lambda_{\scriptscriptstyle M} = D_{d'} = \text{diag}(0,\dots,0,1,\dots,1)$. Consequently, it holds that
    \begin{align}\label{equ:X_as_norm}
        w_n\tr  M w_n = \norm{Mw_n}^2 = \norm{D_{d'} V_{\scriptscriptstyle M}\tr  w_n}^2 = \norm{\tilde{w}_{d'}}^2,
    \end{align}
    where $\tilde{w}_{d'} \in \mathbb{R}^{d'}$. Then, every component of $\tilde{w}_{d'}$ can be written as $\tilde{w}_{d',i} = w_n\tr  v_{{\scriptscriptstyle M}, i}$, where $v_{{\scriptscriptstyle M}, i}$ is an eigenvector of $M$ corresponding to a non-zero eigenvalue.
    By using the law of total expectation and the definition of subgaussianity \cite[Definition 2.2]{wainwright_2019} it holds that
    \begin{align}
        &\BE\left[\exp\left(\lambda\tilde{w}_{d',i}\right)\right] = \notag\\
        &\sum_{a_0 \in \CA}\BE\left[\exp\left(\lambda\tilde{w}_{d',i}\right)\,\vert\, \alpha= a_0\right]\BP\left(\alpha=a_0\right),
    \end{align}
    where $\alpha = [\alpha_1,\dots, \alpha_n]$ {\newtext is the vector of the random signs, see Algorithm \ref{alg:sps_init} (m=2), and $\CA$} is the set of all possible realizations of $\alpha$. Then, by expanding $\tilde{w}_{d',i}$ and using that matrix $M$ only depends on variables $\{\alpha_t\}_{t=1}^n$
    \begin{align}\label{equ:w_di_subgauss}
        &\sum_{a_0 \in \CA}\BE\left[\exp\left(\lambda\tilde{w}_{d',i}\right)\vert \alpha=a_0\right]\BP\left(\alpha=a_0\right)=\notag\\
        &=\sum_{a_0 \in \CA}\BE\left[\exp\left(\lambda  w_n\tr  v_{{\scriptscriptstyle M}, i}\right)\vert \alpha=a_0\right]\BP\left(\alpha=a_0\right)\notag\\
        &=\sum_{a_0 \in \CA}\BE\left[\exp\left(\lambda  \sum_{t=1}^nW_tv_{{\scriptscriptstyle M}, i,t}\right)\Bigg\vert \alpha=a_0\right]\BP\left(\alpha=a_0\right)\notag\\
        &=\sum_{a_0 \in \CA}\BE\left[\prod_{t=1}^n\exp\left(\lambda W_tv_{{\scriptscriptstyle M}, i,t}\right)\Bigg\vert \alpha=a_0\right]\BP\left(\alpha=a_0\right)\notag\\
        &=\sum_{a_0 \in \CA}\prod_{t=1}^n\BE\left[\exp\left(\lambda W_tv_{{\scriptscriptstyle M}, i,t}\right)\Bigg\vert \alpha=a_0\right]\BP\left(\alpha=a_0\right)\notag\\
        &\leq\sum_{a_0 \in \CA}\prod_{t=1}^n\exp\left(\lambda^2 \sigma^2 \left(v_{{\scriptscriptstyle M}, i,t}^{a_0}\right)^2/2\right)\BP\left(\alpha=a_0\right)\notag\\
        &=\sum_{a_0 \in \CA}\exp\left(\frac{\lambda^2 \sigma^2}{2} \sum_{t=1}^n\left(v_{{\scriptscriptstyle M}, i,t}^{a_0}\right)^2\right) \BP\left(\alpha=a_0\right)\notag\\    
        &= \exp\left(\frac{\lambda^2 \sigma^2}{2}\right),\notag
    \end{align}
    where we used the fact that the elements of $w_n$ are independent $\sigma$-subgaussian random variables (\ref{assu:noise}) and that the eigenvectors $v_{{\scriptscriptstyle M}, i}^{a_0}$ are unit length for any realization $a_0$. The norm $\norm{\tilde{w}_{d'}}^2$ can be written as $\sum_{i=1}^{d'} \tilde{w}_{d',i}^2$. It has been shown in \cite[Appendix B]{honorio14} that the square of a $\sigma$-subgaussian random variable is subexponential with parameters $(4\sqrt{2}\sigma^2,4\sigma^2)$. {\newtext Recall that a random variable $Y$ is subexponential \cite[Definition 2.7]{wainwright_2019} if there are non-negative parameters $(\nu,\xi)$ such that
    \begin{equation}\label{equ:subexponential_def}
        \BE\left[e^{\lambda(Y-\BE[Y])}\right]\leq e^{\frac{\lambda^2\nu^2}{2}} \quad \text{for all} \abs{\lambda} < \tfrac{1}{\xi}.
    \end{equation}
    {\roundtwo 
    By using the definition of \eqref{equ:subexponential_def} 
    and introducing the positive scalars $r_1,\dots, r_n$, such that $\sum_{i=1}^n 1/r_i = 1$, Hölder's inequality can be applied to derive that the sum of subexponentials $\{Y_i\}$ with parameters $\{(\nu_i,\xi_i)\}$ satisfies
    \vspace{-1mm}
    \begin{align}
        &\BE\left[e^{\lambda\sum_{i=1}^n(Y_i-\BE[Y_i])}\right] = \BE\left[\bigg|\prod_{i=1}^n e^{\lambda(Y_i-\BE[Y_i])}\bigg|\right]\\
        &\leq\prod_{i=1}^n\left(\BE\left[ e^{r_i\lambda(Y_i-\BE[Y_i])}\right]\right)^{\!1/r_i} \leq  e^{\frac{\lambda^2\sum_{i=1}^nr_i\nu_i^2}{2}},
    \end{align}
    for $\abs{\lambda} < 1/(\max_i\xi_i)$.}
    By choosing $r_i = (\sum_{j=1}^n \nu_j)/\nu_i$,
    \vspace{-1mm}
    \begin{align}
        \BE\left[e^{\lambda\sum_{i=1}^n(Y_i-\BE[Y_i])}\right] \leq e^{\frac{\lambda^2\left(\sum_{i=1}^n\nu_i\right)^2}{2}},
    \end{align}
    thus, the sum of $n$ {\em not necessarily independent} subexponentials with parameters $(\nu_i,\xi_i)$ is subexponential with parameters $(\sum_{i=1}^n\nu_i,\max_i\xi_i)$.}
    From this, it follows that $X = \sum_{i=1}^{d'} \tilde{w}_{d',i}^2 \leq \sum_{i=1}^{d} \tilde{w}_{d',i}^2$ is subexponential with parameters $(4d \sigma^2\sqrt{2},4\sigma^2)$.
    Then, the following inequality holds for $X = w_n\tr  M w_n$ \cite[Proposition 2.9]{wainwright_2019}{\roundtwo :}
    \begin{align}
        &\BP\left(\frac{\vert X - \mathbb{E}X\vert}{n\lambda_0}\hspace{-0.3mm} \geq\hspace{-0.3mm} \varepsilon\right) \hspace{-0.3mm}\leq\hspace{-0.3mm} \notag
        \begin{cases}
            2\exp(-\frac{\varepsilon^2n^2\lambda_0^2}{64d^2\sigma^4}) &\! 0 \leq \varepsilon \leq 8\sigma^2 {d^2}\\[2mm]
            2\exp(-\frac{\varepsilon n\lambda_0}{8\sigma^2}) &\! \varepsilon > 8\sigma^2{d^2}\qed
        \end{cases}\\[-8mm]
        \notag
    \end{align}
\section{Proof of Lemma \ref{lemma:denominator_indicator}}\label{sec:proof_den_indicator}
\vspace{-1mm}
    By expanding $K = \Phi_{{\scriptscriptstyle Q},n}\tr  D_{\alpha,n}\Phi_{{\scriptscriptstyle Q},n}$ \eqref{equ:K_and_L_def}, we get
	\begin{align}
		&\BP\left(\max_i \abs{\lambda_i(\Phi_{{\scriptscriptstyle Q},n}\tr  D_{\alpha,n}\Phi_{{\scriptscriptstyle Q},n})}\geq \varepsilon_0\right) = \notag\\
  		&\BP\left(\max_i \abs{\lambda_i\left(\sum_{t=1}^n \alpha_t\varphi_{{\scriptscriptstyle Q},t}\varphi_{{\scriptscriptstyle Q},t}\tr \right)}\geq \varepsilon_0\right) = \notag\\
        &\BP\left(\max_i \abs{\lambda_i\left(\frac{1}{n}\sum_{t=1}^n n\alpha_t\varphi_{{\scriptscriptstyle Q},t}\varphi_{{\scriptscriptstyle Q},t}\tr \right)}\geq \varepsilon_0\right),
	\end{align}
	where $\varphi_{{\scriptscriptstyle Q},t}$ is the $t$-th row vector of $\Phi_{{\scriptscriptstyle Q},n}$.
    {\newtext Recall that a zero-mean symmetric random matrix $Z \in \BR^{d\times d}$ is subgaussian \cite[Definition 6.6]{wainwright_2019} with a positive semidefinite matrix parameter $V_{\scriptscriptstyle Z}\in \BR^{d\times d}$ if for all $\lambda \in \BR$
    \begin{align}
        \BE[\exp(\lambda Z)] = \sum_{k=0}^{\infty}\frac{\lambda^k}{k!}\BE[Z^k] \preceq \exp\left(\frac{\lambda^2V_{\scriptscriptstyle Z}}{2}\right),
    \end{align}
    where $\BE[\exp(\lambda Z)]$ is the moment generating function of the matrix $Z$.
    Using the above definition of a subgaussian random matrix, in \cite[Example 6.7]{wainwright_2019} it is shown, that if $Z = \zeta Y$, where $\zeta$ is a Rademacher random variable and $Y$ is a deterministic and symmetric matrix, then $Z$ is subgaussian with matrix parameter $V_{\scriptscriptstyle Z} = Y^2$.
    From this result it follows} 
    that $K_t = n\alpha_t\varphi_{{\scriptscriptstyle Q},t}\varphi_{{\scriptscriptstyle Q},t}\tr $ is subgaussian with matrix parameter $V_{K_t} = (n\varphi_{{\scriptscriptstyle Q},t}\varphi_{{\scriptscriptstyle Q},t}\tr)^2,  \forall\, t$, since $\{\alpha_t\}$ are i.i.d. Rademacher random variables and $\varphi_{{\scriptscriptstyle Q},t}\varphi_{{\scriptscriptstyle Q},t}\tr $ is symmetric. {\newtext A Hoeffding bound on random matrices \cite[Theorem 6.15]{wainwright_2019} together with the fact that for symmetric matrices, the spectral norm equals the largest absolute value of the eigenvalues, can be applied to give an upper bound of the above probability as}
	\begin{align}\label{equ:prob_ub_sigmakappa}
		&\BP\left(\lambda_{\text{max}}\left(\frac{1}{n}\sum_{t=1}^n n\alpha_t\varphi_{{\scriptscriptstyle Q},t}\varphi_{{\scriptscriptstyle Q},t}\tr \right)\geq \varepsilon_0\right) \leq \notag\\ &2d\exp\left(-\frac{n\varepsilon_0^2}{2\sigma_{\kappa}^2}\right),
	\end{align}
	where\vspace{-2mm}
	\begin{align}
		\sigma_{\kappa}^2
        &=\max_i \abs{\lambda_i\left(\frac{1}{n}\sum_{t=1}^n V_{K_t}\right)}\notag\\
        &=\max_i \abs{\lambda_i\left(\frac{1}{n}\sum_{t=1}^n \left(n\varphi_{{\scriptscriptstyle Q},t}\varphi_{{\scriptscriptstyle Q},t}\tr \right)^2\right)}\notag\\
        &= \lambda_{\text{max}}\left(\frac{1}{n}\sum_{t=1}^n \left(n\varphi_{{\scriptscriptstyle Q},t}\varphi_{{\scriptscriptstyle Q},t}\tr \right)^2\right).
	\end{align}
    An upper bound on $\sigma_{\kappa}^2$ can be given by using the triangle inequality for the spectral norm (largest eigenvalue),
	\begin{align}	
		  \sigma_{\kappa}^2 &= n\lambda_{\text{max}}\left(\sum_{t=1}^n 	\left(\varphi_{{\scriptscriptstyle Q},t}\varphi_{{\scriptscriptstyle Q},t}\tr \right)^2\right)  \notag\\
		 &\leq n \sum_{t=1}^n \lambda_{\text{max}}\left( \left(\varphi_{{\scriptscriptstyle Q},t}\varphi_{{\scriptscriptstyle Q},t}\tr \right)^2\right) = n \sum_{t=1}^n \norm{\varphi_{{\scriptscriptstyle Q},t}}^4,
	\end{align}
 	where we used that an outer product of a vector $v$ with itself $vv\tr $ has exactly one non-zero eigenvalue which equals $v\tr  v$.
	Using \ref{assu:Phi_Q_coherence} and the fact that $\Phi_{{\scriptscriptstyle Q},n}$ is an orthonormal matrix we conclude that
	\begin{align}
		\sigma_{\kappa}^2 &\leq n \sum_{t=1}^n \norm{\varphi_{{\scriptscriptstyle Q},t}}^4 \leq n \max_t \norm{\varphi_{{\scriptscriptstyle Q},t}}^2 \sum_{t=1}^n \norm{\varphi_{{\scriptscriptstyle Q},t}}^2 \notag\\
		&\leq {\newtext n \, d\kappa n^{-\rho} \norm{\Phi_{{\scriptscriptstyle Q},n}}^2_{\text{F}}} = n \,d\kappa n^{-\rho}\, d = n^{1-\rho}\kappa d^2.
	\end{align}
	Writing back the upper bound on $\sigma_{\kappa}^2$ to \eqref{equ:prob_ub_sigmakappa} we get
	\begin{align}
		\BP\left(\max \abs{\lambda_i(K)} \geq \varepsilon_0\right) &\leq 2d\exp\left(-\frac{n\varepsilon_0^2}{2n^{1-\rho}\kappa d^2}\right) \notag\\
		&=2d\exp\left(-\frac{n^{\rho}\varepsilon_0^2}{2\kappa d^2}\right).\\[-8mm]
        \notag
	\end{align}\qed

\section{Proof of Lemma \ref{lemma:sample_complex_indicator_m2_q1}}\label{sec:proof_sample_complex_indicator_m2_q1}
\vspace{-2mm}
    The results of Lemma \ref{lemma:numerator_indicator} and \eqref{equ:indicator_sec_term_inequ} will be combined to provide the claimed stochastic lower bound.

    Using Lemma \ref{lemma:numerator_indicator}, if $0 \leq \varepsilon \leq 8\sigma^2 d^2$, we have
    \begin{align}\label{equ:lemma2_1}
        \BP\left(\frac{\vert X - \mathbb{E}X\vert}{n\lambda_0} \geq \varepsilon\right) \leq 2\exp\left(-\frac{\varepsilon^2n^2\lambda_0^2}{64d^2\sigma^4}\right).
    \end{align}
    {\newtext By introducing $\delta \defeq 4\exp(-\varepsilon^2n^2\lambda_0^2/(64d^2\sigma^4))$, \eqref{equ:lemma2_1} can be reformulated as, for all $\delta${\roundtwo, such that}
    $4\exp(-(nd\lambda_0)^2) \leq \delta \leq 2$, with probability (w.p.) at least $1-\delta/2$, we have}
    \begin{align}
        \frac{\vert X - \mathbb{E}X\vert}{n\lambda_0} \leq \frac{8d\sigma^2\ln^{\frac{1}{2}}(\tfrac{4}{\delta})}{n\lambda_0}.
    \end{align}
    {\newtext Similarly, if $\varepsilon > 8\sigma^2{d^2}$, for all $\delta: 0 \leq \delta < 4\exp(-(nd\lambda_0)^2) $ it holds, with probability (w.p.) at least $1-\delta/2$, that}
    \begin{align}
        \frac{\vert X - \mathbb{E}X\vert}{n\lambda_0} \leq \frac{8\sigma^2\ln(\tfrac{4}{\delta})}{n\lambda_0}.
    \end{align}
    Putting these together we get, w.p.\ at least $1-\delta/2$, that 
    \begin{align}
        &\frac{\vert X - \mathbb{E}X\vert}{n\lambda_0} \leq 
        \begin{cases}
            \dfrac{8d\sigma^2\ln^{\frac{1}{2}}(\tfrac{4}{\delta})}{n\lambda_0} \vspace{1mm}& {\newtext 4\e^{-(nd\lambda_0)^2} \leq \delta \leq 2},\\[3mm]
            \dfrac{8\sigma^2\ln(\tfrac{4}{\delta})}{n\lambda_0} & {\newtext 0 < \delta < 4\e^{-(nd\lambda_0)^2}}. \\
        \end{cases}
    \end{align}
    In the proof of Lemma \ref{lemma:numerator_indicator} it was shown that $X$ can be upper bounded as $X \leq \sum_{i=1}^d \tilde{w}_{d',i}^2$, where $\{\tilde{w}_{d',i}\}$ are (zero mean) $\sigma$-subgaussian. Then, it holds that
    \begin{align}
        \BE[X] &\,\leq\, \BE\left[\sum_{t=1}^d \tilde{w}_{d',i}^2\right] = \sum_{t=1}^d\BE\left[\tilde{w}_{d',i}^2\right] =
        \sum_{t=1}^d \text{Var}\left[\tilde{w}_{d',i}\right] \notag\\
        &\,\leq\, \sum_{t=1}^d\sigma^2 = d\sigma^2.
    \end{align}
    Using the reverse triangle inequality and the property that $\mathbb{E}[X] \leq d\sigma^2$ we have that 
    \begin{align}
        \frac{\vert X - \mathbb{E}X\vert}{n\lambda_0} \geq \frac{\vert X\vert - \vert\mathbb{E}X\vert}{n\lambda_0} \geq \frac{\vert X\vert}{n\lambda_0} - \frac{d\sigma^2}{n\lambda_0},
    \end{align}
    and it holds w.p. at least $1-\delta/2$ that
    \begin{align}\label{equ:num_result}
        &\frac{\vert X\vert}{n\lambda_0} \leq \notag\\[2mm]
        &\begin{cases}
            \dfrac{d\sigma^2\left(8\ln^{\frac{1}{2}}(\tfrac{4}{\delta})+1\right)}{n\lambda_0}\vspace{1mm} & {\newtext 4\e^{-(nd\lambda_0)^2} \leq \delta \leq 2},\\[3mm]
            \dfrac{\sigma^2\left(8\ln(\tfrac{4}{\delta})+d\right)}{n\lambda_0} & {\newtext 0 < \delta < 4\e^{-(nd\lambda_0)^2}}.
        \end{cases}
    \end{align}
    Now, we reformulate the concentration inequality result of \eqref{equ:indicator_sec_term_inequ} as, w.p. at least $1-\delta/2$:
    \begin{align}\label{equ:den_result}
        \frac{1}{1-\lambda_{\text{max}}(K^2)} \leq \frac{1}{1-\frac{\ln\left(\frac{4d}{\delta}\right)2\kappa d^2}{n^{\rho}}}.
    \end{align}
    By using the union bound it can be derived that, if
    \begin{align}\label{equ:union_1}
        &\BP(Y_1\leq y_1) \geq 1-p_1, &\BP(Y_2\leq y_2) \geq 1-p_2,
    \end{align}
    then
    \begin{align}\label{equ:union_2}
        \BP(Y_1Y_2\leq y_1y_2) \geq 1-(p_1 + p_2).
    \end{align}
    Combining the results of \eqref{equ:union_1}-\eqref{equ:union_2} with the 
    lower bounds of \eqref{equ:num_result} and \eqref{equ:den_result}
    we get that w.p. at least $1-\delta$:
    \begin{align}
        &\frac{\vert X\vert}{n\lambda_0\left(1-\lambda_{\text{max}}(K^2)\right)} \leq \\[3mm]
        &\begin{cases}
            \dfrac{d\sigma^2\left(8\ln^{\frac{1}{2}}(\tfrac{4}{\delta})+1\right)}{n\lambda_0\left(1-\frac{\ln\left(\frac{4d}{\delta}\right)2\kappa d^2}{n^{\rho}}\right)}\vspace{1mm} & {\newtext 4\e^{-(nd\lambda_0)^2} \leq \delta \leq 2},\\[3mm]
            \dfrac{\sigma^2\left(8\ln(\tfrac{4}{\delta})+d\right)}{n\lambda_0\left(1-\frac{\ln\left(\frac{4d}{\delta}\right)2\kappa d^2}{n^{\rho}}\right)} & {\newtext 0 < \delta < 4\e^{-(nd\lambda_0)^2}}. \notag\\
        \end{cases}
    \end{align}
    Next, we derive a concentration inequality for the size of the 0.5-level SPS region utilizing the above stochastic lower bound.
    As we showed in \eqref{equ:main_inequ}, we have for every parameter $\theta \in \confreg[0.5,n]$ the property
    \begin{align}
        \|\tilde{\theta}_c\|^2 &\leq \frac{b\tr  A^{\dagger}b -c}{\lambda_\text{min}(A)} = \frac{\frac{1}{n}w_n\tr  M w_n}{\frac{1}{n}\lambda_\text{min}(A)} \\
        &\leq \frac{\vert X\vert}{n\lambda_0(1-\lambda_{\text{max}}(K^2))}.\notag
    \end{align}
    Therefore, it holds w.p.\ at least $1-\delta$ that\vspace{-1mm}
    \begin{align}\label{equ:theta_c_bound_m2_indicator}
        &\sup_{\theta \in \confreg[0.5,n]}\|\tilde{\theta}_c\| \leq \\[1mm]
        &\begin{cases}
            \left(\dfrac{d\sigma^2\left(8\ln^{\frac{1}{2}}(\tfrac{4}{\delta})+1\right)}{n\lambda_0\left(1-\frac{\ln\left(\frac{4d}{\delta}\right)2\kappa d^2}{n^{\rho}}\right)}\right)^{\frac{1}{2}} \vspace{1mm} & {\newtext 4\e^{-(nd\lambda_0)^2} \leq \delta \leq 2},\\[3mm]
            \left(\dfrac{\sigma^2\left(8\ln(\tfrac{4}{\delta})+d\right)}{n\lambda_0\left(1-\frac{\ln\left(\frac{4d}{\delta}\right)2\kappa d^2}{n^{\rho}}\right)}\right)^{\frac{1}{2}} & {\newtext 0 < \delta < 4\e^{-(nd\lambda_0)^2}}. \notag\\
        \end{cases}
    \end{align}
    By introducing the functions
    \begin{align}\label{equ:f_g_definition}
        &f(\delta) \,\defeq\, \notag
        \begin{cases}
            \sigma\hspace{0.5mm}(\hspace{0.3mm}8\hspace{0.3mm}d\ln^{\frac{1}{2}}(\tfrac{4}{\delta})+d)^{\frac{1}{2}} & {\newtext 4\e^{-(nd\lambda_0)^2} \leq \delta \leq 2},\\[3mm]
            \sigma\left(8\ln(\tfrac{4}{\delta})+d\right)^{\frac{1}{2}} & {\newtext 0 < \delta < 4\e^{-(nd\lambda_0)^2}}, \notag\\
        \end{cases}\\[2mm]
        &g(\delta) \,\defeq\, \ln\left(\tfrac{4d}{\delta}\right)2\kappa d^2,
    \end{align}
        and using the fact that the distance between any two points in the region is less than twice the upper bound presented in \eqref{equ:theta_c_bound_m2_indicator}, we conclude that
    \begin{align}
        &\sup_{\theta_1,\theta_2 \in \confreg[0.5,n]}\hspace{-4mm}\|\theta_1-\theta_2\| \leq
        \dfrac{2f(\delta)}{\left(n^{1-\rho}\lambda_0\left(n^{\rho}-g(\delta)\right)\right)^{1/2}} \hspace{0.5mm}.
    \end{align}\qed
\end{document}